# What makes a good BIM design: quantitative linking between design behavior and quality

Xiang-Rui Ni[1], Peng Pan[1,2], Jia-Rui Lin[1,2,*]

1 Department of Civil Engineering, Tsinghua University, Beijing 100084, China
2 Key Laboratory of Digital Construction and Digital Twin, Ministry of Housing and Urban-Rural Development, Beijing 100084, China

## Abstract

In the Architecture Engineering & Construction (AEC) industry, how design behaviors impact design quality remains unclear. This study proposes a novel approach, which, for the first time, identifies and quantitatively describes the relationship between design behaviors and quality of design based on Building Information Modeling (BIM). Real-time collection and log mining are integrated to collect raw data of design behaviors. Feature engineering and various machine learning models are then utilized for quantitative modeling and interpretation. Results confirm an existing quantifiable relationship which can be learned by various models. The best-performing model using Extremely Random Trees achieved an $R^2$ value of 0.88 on the test set. Behavioral features related to designer's skill level and changes of design intentions are identified to have significant impacts on design quality. These findings deepen our understanding of the design process and help forming BIM designs with better quality.

Keywords：BIM, Design quality, Design behavior, Machine learning, Model interpretation

## 1 Introduction

In the AEC industry, construction projects are mainly divided into several stages, including design, construction, and operation. Although the design stage accounts for a relatively small amount of time, funding, and manpower in the entire project, the quality of design often plays a decisive role in the overall quality and efficiency of the project [1]. Therefore, ensuring the quality of the design is of significant importance in ensuring the quality and efficiency of the entire project [2]. To ensure design quality, common practice and related studies mainly focus on discovering [3] and resolving [4] design problems. But very little attention is paid on exploring the cause of design problems to reduce the number of them.

During the design stage, the main characters carrying out design activities are designers. Factors such as their proficiency in using design tools, the richness of design experience, and their level of engagement during the design process may directly affect the design behaviors they take [5]. In the relevant research field as well as within this study, design behaviors refer to all relevant actions and operations taken by the designer when working in a BIM environment for specific design objectives. The final design is the accumulation of all these behaviors, therefore the behavioral characteristics of the designers and the quality of the final design results are believed to be inherently interlinked [6].

---

* Corresponding author.
E-mail address: lin611@tsinghua.edu.cn (Jia-Rui Lin)

In engineering practice, people have empirically summarized certain good practices and bad practices in the BIM design process [7,8]. Good practices are considered to have a positive impact on the efficiency and quality of design, whereas bad practices have the opposite effect. In the academic field, qualitative studies have identified that multiple factors, such as individual capabilities, work environment, and design intentions, influence the behaviors adopted by designers, thereby affecting the quality of the design [9]. However, qualitative studies have their limitations. Some scholars have pointed out that even for experienced designers, data-driven approaches can still provide new insights, better design strategies and optimized design processes, thereby enhancing the quality and efficiency of design [10]. Establishing a data-driven quantitative model that intuitively represents the relationship between design behaviors and design quality can take a step further compared to these qualitative studies. It helps in understanding the intrinsic reasons for variations in design quality, validate from a data perspective whether certain empirically derived practices are truly effective, and potentially offer more effective guidelines for design behaviors. This, in turn, may assist us in fundamentally improving design quality and efficiency, and reducing the occurrence of design errors.

Currently, very few studies focus on analyzing how exactly design behavior results in certain quality of design. Existing studies on the analysis of design behavior data are primarily focused on the identification and prediction of local features instead of overall ones. For example, measuring the efficiency of performing specific operation, using Long Short-Term Memory Network (LSTM) to learn design behavior sequence to predict possible design behavior and using behavior data to analyze collaborative relationship between designers and teamwork efficiency [11]. There remains a notable research gap regarding exploring the relationship between design behavior characteristics and design quality, especially quantitatively modeling the link between the two.

This study aims to extract behavior features from BIM-based design behavior data and subsequently establish quantitative models that proof and describe the relationship between these behavior features and the design quality. By analyzing and interpreting the key features of the developed models, this study seeks to deepen the understanding of design behaviors and provide data-driven insights to form guidance for assuring design quality. The remainder of this paper is organized in the following structure. Section 2 reviews existing studies on design quality management, quality analysis using behavior data in other disciplines as well as in the AEC industry and existing interpretation method for machine learning. Section 3 provide the methodology adopted by the study in full details including methods for data collection, data preprocessing, data augmentation, feature selection, modeling experiments and model interpretation. Section 4 illustrates the collected data set and analyzes the results of the experiments and proposes the optimal model and the results of model interpretation. With these results, section 5 discusses the significance of the study, potential value of reference for other related studies, the limitations of this study and possible directions for future research. Finally, Section 6 concludes this study.

# 2 Literature Review

## 2.1 BIM-based Design Quality Assurance

Design quality issues have long been one of the critical challenges in engineering practice. Even well-developed designs require rigorous reviews before construction stage to identify and resolve design problems [12]. As a key information technology in the AEC industry, BIM provides an integrated platform for managing project whole life cycle data, greatly facilitating communication and collaboration among project stakeholders [13]. Moreover, the three-dimensional modeling environment offered by BIM is more conducive to identifying design issues, particularly those related to multi-disciplinary coordination, and helps designers intuitively investigate potential solutions.

The main categories of BIM design quality issues include [14]: 1) Completeness: missing necessary components in the model or lacking required property information; 2) Correctness: dimensions or property information of model components not aligning with design knowledge and requirements, or improper spatial relationships between components; 3) Compliance: model components or parameters not meeting relevant design standards; 4) Consistency: inconsistency between drawings and models, inconsistent data representing the same information stored in different locations, or data inconsistencies related to geometry and other information; 5) Multi-disciplinary coordination: collisions between components from different disciplines, insufficient installation space, or inadequate clearance in the integrated model.

Current industry practice for quality assurance primarily relies on manual identification and discussion to resolve these issues, which is inefficient and risks overlooking problems [15]. Very few of these issues can be addressed using automated methods. Existing studies have explored more effective automated approaches for identifying and resolving these design quality problems based on BIM models. For completeness issues, studies have proposed rule-based identification and machine learning or knowledge graph-based methods for missing data imputation [16,17]. For compliance issues, researchers have extracted information from standard texts and converted them into machine-executable structured formats to enable automated BIM design compliance checking [18].

However, the existing studies have primarily focused on checking and resolving identified design problems, with little research addressing the root causes of these issues from the designers' perspective and how to more directly improve design quality and reduce their occurrence.

## 2.2 Quality Analysis using Behavior Data

In other disciplines, many studies focused on explaining or predicting the quality by establishing a quantitative relationship between process behavior data and quality of the result. Nikola et al. [19] collected the behavior data of students doing online exams and their basic characteristics data. Data of the previous students was then used to predict whether a student can pass the exam. Sven et al. [20] used a model based on Extreme Gradient Boosting algorithm to learn underlying feature patterns of the behavior data of workers in the warehouse. This model can be used to predict the work efficiency of newly recruited employees and make better work arrangements for the warehouse.

In the AEC domain, research on design behavior are primarily categorized into qualitative and quantitative approaches. Qualitative studies have explored the relationship between design behaviors and design quality, typically employing methods such as manual evaluation and surveys to provide limited descriptions of design behaviors. Vegard et al. [21], based on case analyses and interviews, investigated 10 key factors in the architectural design management process, including planning and designer management. Brown [10] designed comparative experiments and, through surveys, analyzed designers' performance under the influence of various factors such as different environments and tools, as well as the quality of the final design. The author also stressed the importance of data-driven approaches which can offer new insights, better design strategies even for experienced designers.

Quantitative methods represent a relatively new field, with fewer studies in this area [22]. They mainly focused on the log files automatically generated by software as the only data source, and the identification or prediction of local features of behavioral data, such as time consumption of specific commands, sequential pattern of commands [23]. There is a lack of relevant studies exploring the direct relationship between design behavior and quality of the results combined with a quality evaluation system. Yarmohammadi et al. [24] extracted design behavior features from the log files automatically generated by BIM modeling software and used the Generalized Suffix Trees algorithm to analyze the behavior of designers executing modeling commands. They found that different designers had significantly different efficiency of executing same series of commands. This may serve as a method to evaluate the working efficiency of different individuals. Pan et al. [25] utilized the integration of the Kohonen Clustering Network (KCN) and Fuzzy C-Means (FCM) to analyze the design efficiency of designers across different time periods. By clustering designers based on this analysis, they aimed to optimize work schedules and enhance team productivity. Zhang et al. [26] used pattern recognition algorithms to obtain the most frequently executed sequences of modeling commands from log data, pointing out that different designers were significantly different in the efficiency of executing specific sequences. In the next year, they also analyzed the log data of different designers working together in the same project to analyze the value of different designers and the social network [27]. Pan et al. [11] extracted command execution data from the log file generated by Autodesk Revit and used the Long Short-Term Memory Network to learn sequences of modeling commands and predict what command is likely to come next.

In summary, while there is a considerable body of research in other academic fields investigating the quantitative relationship between behaviors and quality of results, similar studies within the AEC domain remain relatively scarce. The primary research gaps are: 1) the quantitative relationship between design behaviors and design quality is critically important yet underrepresented, 2) existing quantitative studies often lack a focus on overall characteristics and design quality.

## 2.3 Interpretation Method for Machine Learning Models

This study focuses on establishing a quantitative regression model to describe the relationship between design behavior and design quality. But obtaining a model is not the ultimate goal. This study also attempts to extract some knowledge from the data and models, so as to deepen our understanding of the practical problems of design, and to provide some insights for solving such problems [28]. Therefore, the interpretation of machine learning models is very important, because it helps us understand how the models perform, and really gain some knowledge from the data and models.

For linear model, we can easily understand the structure and the mechanism of the model and we can know every detail of how the input produces the output. For some more complex but more promising non-linear models, the mechanism is usually referred to as a "black box" and it is difficult if not impossible to understand what exactly happened in the model or why the model gives certain results [29,30]. This leads to acquiring an effective model but without any further knowledge or insights.

For some specific models, such as models based on deep learning algorithms and tree-based algorithms, many studies have proposed targeted interpretation methods. In 2017, Scott et al. [31] proposed the general model interpretation method based on Additive feature attribution methods and SHapley Additive exPlanations (SHAP) values, combining six previously proposed model-specific methods. This method has been widely used in recent years in many studies using complex machine learning algorithm-based models.

# 3 Methodology

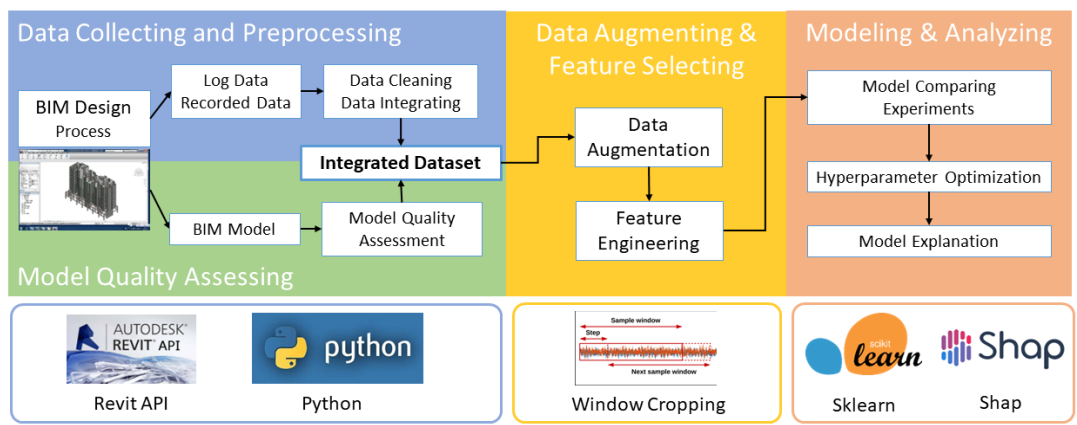

Figure 1. Methodology of this study

As shown in Figure 1, this study first collected behavioral data from the BIM modeling design process based on Autodesk Revit software. Collected data mainly included log data automatically generated by the software and behavioral data collected in real time. The raw data then went through data preprocessing for cleaning low quality data and integrating data collected in different ways. On the other hand, a targeted modeling quality evaluation method was proposed and the modeling results, namely the BIM models, were scored as a quantitative index to represent the design quality. The integrated data went through further data augmentation to ensure the sufficient data size for model training. Feature engineering was utilized to manually select data features for subsequent modeling and analyzing. The modeling and analyzing contained two sets of experiments: 1) model selection experiments to compare the effects of different machine learning algorithms, 2) hyperparameters optimization experiments using controlled variables method to determine the reasonable values of the hyperparameters introduced by the data augmentation method. Finally, the machine learning model interpretation method based on SHAP value was adopted to explain and analyze the model selected by the experiments and obtain important information such as key features of the model providing further insights for improving design quality in practice.

## 3.1 Design Behavior Data Collection and Preprocessing

### 3.1.1 BIM Design Behavior Data Collection

Design behavior is the behavior made by the designer that contributes to the formation of the final design result and the design behavior data contains information needed to comprehensively describe the design behavior, including the type of behavior, the specific behavior information, the time of the behavior and the effect of the behavior. As mentioned before, most relevant studies only used the log data automatically generated by software as the data describing the design behavior for analysis, but there are two obvious deficiencies. 1) The data in log files may not be sufficient. For example, log files produced by Revit lack some of the keyboard input information. 2) Log files are not particularly created for research purposes, therefore considerate amount of effort is required to manually recognize and extract needed information. For example, information about the creation and modification of model components in the Revit log data is incomplete and scattered. Much of it uses specially defined symbols or codes to record data, which makes it even more difficult to interpret [32].

In this study, a novel method was adopted which combines log files and real-time collection method [33]. By developing Revit plugin tools, this method can directly collect data like keyboard pressing, elements created, modified or deleted. The data collected in real time can serve as an important supplement for log files, which can improve the efficiency of extracting log file data and further ensure the quality and integrity of the data. A Revit plugin called Design Tracker was developed based on the Revit open Application Programming Interface (API) [33]. It subscribes to specific events to capture certain behaviors like the designer's modification to the model and the input of the keyboard, and then output related data to the external data files. Considering the data privacy concerns of students, no sensitive or private personal data was collected. Data collected via the Design Tracker was saved in CSV format and log data were in their original TXT format. They were stored in two separate data files and they paired up as complete data describing the design process in one Revit session.

The data source of this study was found in the author's university. Four courses were identified to be focusing on the principles of BIM and modeling methodology based on Autodesk Revit software. These courses included Revit modeling assignments to evaluate the students' learning outcome. The assignments required them to complete a small BIM model based on the given examples. They could choose other models as reference as long as they had similar size and functionality as the examples. They were also allowed to make unique modifications and innovative designs. This study utilized the process of students completing these modeling assignments as the data source. Specifically, students were guided to install the Design Tracker for data collection. As they progressed with their assignments, the plugin recorded their behaviors and saved the data in CSV format in a specific data folder. At the end of each Revit session, the corresponding log file was automatically located and copied to the folder. The students were then required to submit the data folder along with their assignment deliverables after finishing the assignments.

The final collected raw data amounted to 2.21GB, including 3,074 data files from 68 students. Each student submitted an architectural model and a structural model, so a total of 136 BIM models were collected. On average, each model corresponded to about 22 files which can describe about 11 complete Revit sessions. Among them, 29 students had prior practical work experience in the field of civil engineering design before participating in the course, while the remaining students were essentially engaging in 3D modeling for the first time. Therefore, although the data for this study came from students, it still encompassed designers with varying skill levels. The conclusions derived from the analysis of this data are applicable to designers at different proficiency levels.

### 3.1.2 Data Cleaning

The collected raw data needed to be cleaned first to ensure the quality of the data. Log data is automatically generated by Revit software, and the quality is relatively stable; but the data collected by the Design Tracker can be affected by unexpected behavior or programme errors, resulting in missing or mistaken data. We did receive reports from students of unexpected error thrown by the plugin during modeling. These errors caused occasional blank lines and missing command name in the data. A few files exhibited a significantly shorter coverage duration than their corresponding log files due to unexpected early termination of the recording process. These incomplete or incorrect data were removed during the data cleaning stage.

Secondly, the length of data files varied, depending on how much design behavior the user performed in each Revit session. Each row in the data file corresponds to one specific operation behavior. Too short data files may contain little information regarding designer's behavioral characteristics. Therefore, data with too small length were removed in data cleaning. Paired data was discarded in this study if the log data file had a size less than 100 KB.

Overall, removed data and data files constituted only a small fraction of the total data volume, ensuring data quality while not causing any appreciable change in the overall data quantity.

### 3.1.3 Data Integration

After the data cleaning, paired data files used to describe a complete Revit session needed to be merged into one data file to obtain an integrated data set. Data integration depended on the time stamp information of each row in the two files. Rows from the two files were rearranged in time order and output into one data file.

Data integration mainly includes three steps: filtering, formatting and integrating of each specific data record. The process of data integration and a typical example are shown in Figure 2. Data in log files need to be filtered since log files are automatically generated by Revit and contain a significant amount of information that is not related to this study, such as operating system configuration, software configuration information, etc. In this study, out of the total 28,786,373 lines of log data, only 13% were considered relevant. It is worth noting that low data relevance rate is a common phenomenon for BIM log mining studies [32]. For example, Zhang's study [26] extracted only approximately 200MB of relevant data from 4GB of log files, resulting a data relevance rate of only about 5%.

Usually, log record related to the designer's operation contains "Jrn" or "jrn" strings at the beginning of the Data Type, this study filtered related records using this feature. A typical log data contains Time Stamp, Data Type and Data Details. Time stamp is the time of this data record, and its format is cultural dependent. Data Type describes the type of data. In this example, "Command" means that this data is about a command and "AccelKey" means that the command is carried out using shortcut key combination. The Data Type is usually followed by multiple Data Details containing more information related to this data. In this example, the only Data Detail states that the functionality of this command is canceling the current operation.

The data collected in real time has been formatted by the plugin, which has a similar structure as the log files. Each line represents one designer operation and different parts are separated by commas. Each line also contains an extra part called Time Stamp in Ticks. Time expressed in Ticks is the total number of milliseconds that have passed from the time 01-Jan-0001 00:00:00 to the specified current time. This method provides a unified way to express time and makes Time Stamps easily comparable. During formatting, Time Stamps in log files were also converted into time expressed by ticks for time alignment comparison.

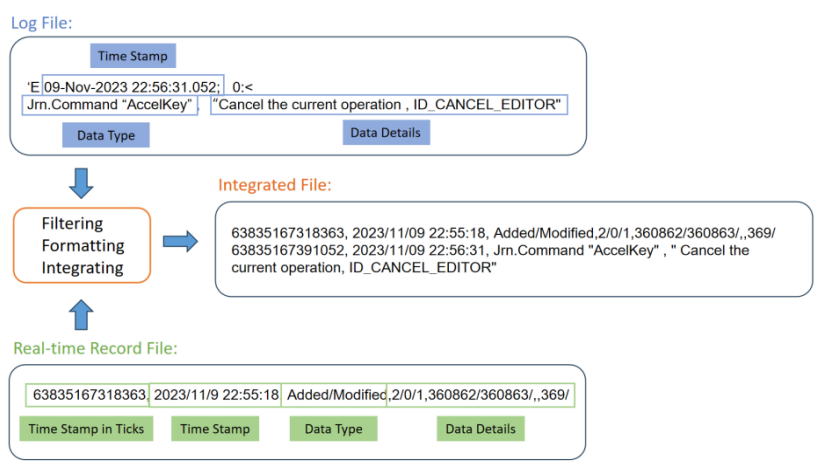

Figure 2. Data integration process

## 3.2 BIM Model Quality Evaluation

Evaluation of BIM design quality is a complex problem. Under different application scenarios, criteria and indicators for evaluation can be different, so it is difficult to find a general method for BIM model evaluation. Some studies suggested a few basic principles that should be considered during evaluation [14]. Existing studies mainly considered completeness, accuracy and multi-professional consistency as three main aspect for model quality evaluation [34].

In this study, model design quality was determined by scoring the students' modeling results with an evaluation method. Considering the characteristics of students and the goal of these courses, general evaluation methods in AEC industry is not suitable to use. Therefore, a targeted evaluation method is needed. The data source for this study was students, most of whom had insufficient modeling capabilities. When specifying the evaluation method, we considered the actual teaching focus and the students' skill level. In actual teaching, due to the limitations of course content volume, the introduction to architectural models was relatively detailed, with more teaching and practice of various modeling functions, while the teaching related to structural models was more concise and preliminary. Moreover, not every student possessed sufficient knowledge to design a structural model that adheres to the principles of mechanics and meets engineering standards.

After consulting with the course instructors and experts, the evaluation of student model quality in this study was divided into four parts for architectural and structural models submitted by students. For architectural models, three indicators — completeness, accuracy, and complexity—were assessed, each carrying equal weight, with an initial score of 20 points each. For structural models, only completeness was evaluated. The requirements for structural modeling assignment were only creating certain types of structural components in reasonable positions, such as foundation, rebar, etc., and did not require their size, number or position to be correct and in line with structural engineering standards. The weight for structural completeness was halved, with an initial score of 10 points. The total initial score was 70 points. The sum of scores for the architectural model and the structural model was the total score for one student's modeling result, which was used as the quantitative index for the evaluation of BIM design results.

Scores were given in a comparing way. One student's modeling results was firstly selected as the benchmark, and all aspects were scored with basic score which sum to a total score of 70 points. The scores of all other students were obtained by comparing with the benchmark, awarding extra points or deducting points in each aspect.

For evaluating the architectural model, three aspects were considered: completeness, accuracy and complexity. Completeness measured whether students had created all the required types of building components with reasonable placement such as walls, columns, floors, doors, doors, Windows, stairs, etc. Missing types of components caused deduction of the evaluation score. Accuracy measured whether there were visible modeling errors in the model, such as wall members mistakenly passing through the roof, missing necessary platform slabs in the stairwell. More mistakes resulted in lower score. Complexity measured the scale and complexity of the model. The larger the scale, the more types of components included, the more complex the components were, the higher the score. Each missing type of components deducted 2 points, each modelling error deducted 2 points. Awarding or deducting points of complexity evaluation shall not exceed ±10 points.

For evaluating the structural model, the only aspect considered was completeness which has the same meaning described in the last paragraph. Each missing type of components or unreasonable placement deducted 2 points. The first part of the scoring sheet for this study is shown in Figure 3.

Scoring Sheet

| Student Number | Architecture Model | | | Structural Model | Total Score |
|---|---|---|---|---|---|
| | Completeness | Accuracy | Complexity | Completeness | |
| 1 | 20 | 20 | 20 | 10 | 70 |
| 2 | 20 | 18 | 22 | 8 | 68 |
| 3 | 20 | **18** | **28** | 10 | 78 |
| ... | | | | | |

Row 1: Benchmark Score
-2 Deducting Points for one extra mistake (Accuracy, Student 3)
+8 Awarding Points for extra complexity (Complexity, Student 3)

Figure 3. Model scoring

Based on the advice of the course instructors, we selected a student with an average performance as the benchmark and assigned every part of his score as the initial score. Based on the performance of this student, we estimated the possible ranges of scores for each part and in total. For example, the range of the completeness score of architectural models was estimated based on the following considerations. The architectural model components we demanded included six categories: walls, doors, windows, floors, stairs and roofs. The roofs and stairs of the selected student's architectural model were incomplete, so the best performance for this part was to fully create these two types of components on this basis, with a maximum score of 24 points. Considering that in most cases, there should be at least one type of components meeting the requirements, so the worst-case scenario would be failing in 3 more types, with a score of 14 points.

For architectural models, the completeness score should range from 14 to 24. The maximum value of the accuracy score should be 24, and the minimum value was estimated to be around 12. According to the evaluation method, the complexity score naturally ranged from 10 to 30. For structural models, the integrity score should range from 2 to 16. The total score should be between 38 and 94. It should be pointed out that these are rough estimations that considered extreme cases which the actual data may include very few if not none. The actual ranges of students' scores are smaller than these estimations.

We invited two experts in the field of BIM from the author's affiliated institution. They possessed necessary knowledge in BIM design and modeling as well as engineering experience, as they had been deeply involved in many engineering projects. They were asked to score all students modeling results according to the above method. It is worth pointing out that, although this study had clearly defined the criteria for scoring, the scores were still given by people, which inevitably had some subjectivity. This is an important feature of the data set in this study, and section 4 will analyze the impact of this feature on data analysis.

### 3.3 Data Augmentation and Feature Engineering

### 3.3.1 Data Augmentation

In this study, students from related courses were considered as the data source so that the number of samples was relatively small. Therefore, in order to make the regression model more fully trained, data augmentation was necessary to ensure a sufficient amount of data. Data is recorded in rows or in the form of sequence and there are many augmentation methods for sequence-type data. Window cropping method [35] was adopted for this study. This method requires setting a cropping window moving along the sequence data. Each data file is sliced into a number of sub-sequences with a length of $N$, and the cropping window moves forward with a step size $s$. The size $n$ of the overlapping part between two adjacent sub-sequences can be calculated as: $n = N - s$. The corresponding quality score for each newly generated data sample is consistent with the original data.

On one hand, this data augmentation method can balance the model learning bias caused by different sizes of samples, on the other hand, it can reduce the impact of noise data on the model and improve the robustness and generalization capability of the model.

### 3.3.2 Feature Engineering

In this study, the features extracted through feature engineering are mainly the overall densities of selected types of records in the samples obtained through statistical methods, which are defined into two categories: data density and time density. Data density refers to the proportion of the number of specified type of records to the total number of records in the sample, while time density refers to the proportion of the total time consumed by such records to the total time span of the sample. The time consumed by a record is calculated as the Time Stamp difference between this record and the previous one. For analysis purpose, the indications of the two categories are different. Data density may reflect the designer's behavioral preferences, while the time density may reflect the efficiency-related information of the designer.

To calculate the density, the total length ($L$) and the total time span of the sample ($T$) is first calculated. Table 11 presents direct sample statistical indicators and abbreviations for features obtained from them. Subsequent analysis will be using these abbreviations. The selection of statistical indicators mainly considered their possible role in reflecting the designer's effective design behavior and behavioral characteristics, as well as the availability of information in the integrated data set.

Table 11. Statistical indicators and features.

| No. | Statistical Indicators | Data Density Abbreviations | Time Density Abbreviations |
| --- | --- | --- | --- |
| 1 | Number of records with "Transaction Successful" | transsuccess% | transsuccess%t |
| 2 | Number of added components | add% | add%t |
| 3 | Number of adding | add_times% | add_times%t |
| 4 | Number of deleted components | delete% | delete%t |
| 5 | Number of deleting | delete_times% | delete_times%t |
| 6 | Number of modifying | modify_times% | modify_times%t |
| 7 | Number of executed commands | command% | command%t |
| 8 | Number of undo or cancel commands | undo% | undo%t |
| 9 | Number of commands via ribbon buttons | ribbon% | ribbon%t |
| 10 | Number of command via shortcuts | accelkey% | accelkey%t |
| 11 | Number of pushing command related buttons | pushbutton% | pushbutton%t |
| 12 | Number of pauses of 1-2 minutes | idle1-2% | idle1-2%t |
| 13 | Number of pauses of 2-5 minutes | idle2-5% | idle2-5%t |
| 14 | Number of pauses of more than 5 minutes | idle>5% | idle>5%t |
| 15 | Effective work time | - | effectT% |

The reasons of selecting these statistical indicators are as follows.

1) **No.1, Records with "Transaction Successful":** Records with the phrase "Transaction Successful" come from log files, which usually represent a group of modifications of the BIM model have been successfully executed and recorded through Revit's own transaction mechanism. Therefore, the density of such records can reflect how many substantial modifications the designer performed to the model, which may imply the working efficiency of the designer.

2) **No. 2-6, Records related to component manipulation:** Records of adding, deleting and modifying elements come from data files collected via the Design Tracker, which can also reflect the efficiency of the designer for different types of specific modification behaviors. Among them, the total number of modified elements is not considered as a valuable statistical information. This is because in the BIM model components usually have many dependency relationships and geometry constraints with other components. When these components are modified, the software will automatically modify every other component which has relationship with them to ensure that dependency, constraint and other relationships remain valid and correct. Usually, the number of components actually modified is much larger than the number of components that the designer really cares about. Therefore, "the total number of modified elements" may not reflect the designer's behavior characteristics or intention and therefore is not considered as a valuable statistical indicator in this study.

    3) **No.7-11, Records related to commands:** In the BIM-based model design process, the most important behavior is executing commands. Many statistical indicators related to the execution of commands are considered important, including the total number of commands executed, the number of commands canceled or undone, the number of commands executed through the ribbon menu, the number of commands executed in shortcuts and the number of buttons pressed. These indicators mainly reflect the execution of commands and the way of execution. Of course, commands can also be divided into more specific types according to their specific function, but considering the excessive variety of commands and the number of executions of some rarely used commands may be too small to be significant, this study does not make more detailed classification of commands.

    4) **No.12-14, Records related to pauses:** Designers may turn to deal with other affairs or take a rest in the middle of the design process, resulting in a pause in the data sample, that is, an interval of time between the two adjacent data records. In general, considering the time of the operation itself and the response time of the computer, the time interval between continuous data records is in the range of milliseconds to seconds. There are also some specific commands that are rarely used and require long computing time, which may take several minutes. Therefore, pauses of more than 1 minute in the data samples are usually not due to the limitation of computing power but the designers consciously delaying or halting the ongoing design process. Designers may be engaged in contemplation or momentary interruptions in case of short pauses and dealing with other tasks or leaving the work place in case of long pauses. The number and frequency of short pauses may reflect the designer's design efficiency, concentration level at that time. In this study, time intervals shorter than 1 minute were not counted as pauses, and the intervals of 1-2 minutes, 2-5 minutes and over 5 minutes were calculated as three different categories for different lengths of time intervals may imply different causes of pauses.

    5) **No.15, Effective work time:** The last time-related feature is calculated from the statistical indicator "effective work time". In this study, effective work time is defined as the sum of the total time span of the sample excluding the sum of all time intervals over 5 minutes. A pause of more than 5 minutes usually means the designer is no longer working on the design for a significant long time, so this period is not counted as work time. Shorter time intervals implies that the designer does not completely halt the current work, so this part of the time is still considered as working time only with less efficiency. Dividing the effective work time by the total time span of the sample is the time density of the effective time, which may reflect the continuity of the designer's behavior and the designer's depth of engagement. The data density calculated with the effective work time is not of obvious practical significance and therefore is not used as feature.

    The number of manually extracted features in this study totals 29. The number of features is small, so methods such as correlation matrix analysis are no longer used for feature selection. Of

course, this number of features still has demand for the size of the data set used for the subsequent machine learning model training, and data augmentation should produce a sufficient number of samples.

## 3.4 Modeling Experiments and Analysis Methods

### 3.4.1 Model Selection Experiments

Model selection experiments were used to identify the most effective supervised machine learning algorithms for establishing quantitative models. The data set used in the experiment was the behavioral feature data set obtained after the data augmentation and feature engineering. The data set was randomly split in an 80-20 ratio, with 80% of the data allocated to the training set and the remaining 20% allocated to the test set. Considering the two hyperparameters introduced in the data augmentation step, during the model selection experimental stage, all the algorithms were trained with an identical set of hyperparameters configured based on the characteristics of the data set.

The experiments used the Sklearn software package of Python to iterate through all the available supervised machine learning regression algorithm for modeling training. The expected output of the model was the score corresponding to the sample, and the mean square error of the output score was taken as the loss function. The root mean square error (RMSE) and Coefficient of Determination ($R^2$) value were used to evaluate the effect of the trained model on the test set. The most effective models were then used in subsequent hyperparameter optimization experiments to determine the appropriate hyperparameter values.

Sklearn package provides a number of available algorithms and the following categories were considered as candidates in this study: Linear Algorithms (including Ordinary Least Squares, Ridge regression, Lasso, Elastic-Net, etc.), Support Vector Machine, Stochastic Gradient Descent, Nearest Neighbors, Gaussian Processes, Decision Tree Based and Ensemble Methods (including AdaBoost，Bagging，Random Forest, etc.).

The use of both RMSE and $R^2$ provides a well-rounded assessment of a model's performance and effectiveness. RMSE is used to describe the average error between the score value output by the model and the true score value. The smaller the RMSE value is, the better the model fits. Given a test set $Q$, which contains $n$ samples, the true rating for each sample $i$ is denoted as $y_i$, and the model's evaluated rating is denoted as $\hat{y}_i$. The RMSE of the model on the test set can be calculated as equation (1).

$$RMSE = \sqrt{\frac{1}{n}\sum_{i=1}^{n}(\hat{y}_i - y_i)^2} \qquad (1)$$

The $R^2$ value is a metric that measures the proportion of the variance in the dependent variable that is explained by the model from the independent variables. It represents the extent to which the model can explain the variability in the target variable. Under the same assumptions as before, the $R^2$ value of the model on the test set $Q$ can be calculated as equation (2):

$$R^2 = 1 - \frac{\sum_{i=1}^{n}(y_i - \hat{y}_i)^2}{\sum_{i=1}^{n}(y_i - \bar{y})^2}, \quad where \quad \bar{y} = \frac{1}{n}\sum_{i=1}^{n} y_i \qquad (2)$$

The $R^2$ typically ranges from 0 to 1. An $R^2$ of 1 indicates that the model can completely explain the variability in the data, while an $R^2$ of 0 suggests the model has no explanatory power and cannot account for any of the variability in the target variable.

In the case where the errors in the model's outputs exhibit greater variance than the original data, the $R^2$ may become negative. This situation implies that the model performs even worse than simply using the mean value to explain the data. Since the mean value represents none of the variance, this specific model is not an appropriate choice and should not be employed, as it fails to provide any meaningful interpretation of the data's variability.

### 3.4.2 Hyperparameter Optimization

The data augmentation method introduced two hyperparameters in this study, the length *N* and the step *s*. Different hyperparameter values can affect the total number of samples, the length of each sample and the similarity between samples. Different training data can results in models of different effectiveness. Therefore, it is necessary to determine a set of appropriate hyperparameter values through experiments.

The experiments adopted the controlled variable method, fixing one parameter while changing another parameter to observe the effect of change on the same model.

### 3.4.3 Model Interpretation Method

To identify the most salient behavioral features and explain model performance, this study employed a SHAP value-based approach to conduct an interpretation analysis on the optimal model obtained from the previous experiments. For each sample, a SHAP value can be computed for every feature, which quantifies the contribution of that feature to the model's output for the given input. The SHAP value represents the magnitude of influence that a particular feature has among all the features, in terms of its impact on the model's output.

For data set *Q*, let *N* represent the complete set of all features. Through the training process, we have obtained a regression model *f*. To compute the classical SHAP value $\phi_i(f)$ for feature *i* with respect to the model output, equation (3) is employed:

$$\phi_i(f) = \sum_{S \subseteq N \setminus \{i\}} \frac{|S|!(|N|-|S|-1)!}{|N|!} [f_{S \cup \{i\}}(x_{S \cup \{i\}}) - f_S(x_S)] \tag{3}$$

Where *S* represents a subset of the feature set *N* excluding feature *i*. $f_{S \cup \{i\}}$ and $f_S$ denote additional interpretation models trained on the feature space $S \cup \{i\}$ and *S* respectively. To compute the SHAP value, we must iterate through all possible subsets *S*, assign a weight to each subset, and then sum the weighted contributions to obtain the SHAP value for the feature *i*.

Precisely calculating the SHAP values can be quite challenging, especially when the number of features is large. Therefore, various approximation methods are commonly employed to estimate the SHAP values, while maintaining accuracy and consistency, in order to achieve higher computation efficiency [31].

A positive SHAP value indicates that an increase in the feature's value leads to an increase in the model's output; the larger the SHAP value, the more the model's output increases. Conversely, a negative SHAP value implies that an increase in the feature's value results in a decrease in the model's output; in this case, the larger the absolute value of the SHAP value, the more the model's output decreases. By averaging the SHAP values of a particular feature across all samples, we can assess the overall impact of that feature on the model's output. This approach allows us to identify the most influential features that have the greatest effect on the model's output.

For a trained model, we can utilize the Shap package in Python to compute the SHAP values for each feature, thereby quantifying their contributions to the model's outputs. By analyzing the

SHAP values of the behavioral features, we can identify the most important and influential features. Combining this information with domain-specific knowledge, we can then provide meaningful interpretations of these key features. This approach allows us to gain valuable insights that can inform and inspire solutions to real-world problems. The SHAP values serve as a powerful tool for model interpretation, enabling us to understand the underlying drivers of the model's output and to extract actionable insights from the trained model.

# 4 Results

## 4.1 Integrated Data Set

The raw data obtained from the students' submitted assignment files were consolidated into a unified data set after data cleaning and integration. The resulting integrated data set contains design behavior data from 68 students, with each student's data comprising multiple data files, each representing a complete Revit session. There is a significant variation in the total data volume across different students, as shown in Figure 4. The average total data length is 58,045 rows, with the maximum length approaching 350,000 rows. This disparity in data length highlights the necessity of using a window cropping approach during data augmentation to mitigate the impact of varying data lengths.

Among all the samples, two-thirds of the total data lengths exceed 30,000 rows. This data point serves as the initial value for the cropping length hyperparameter ($N$) in the subsequent experiments.

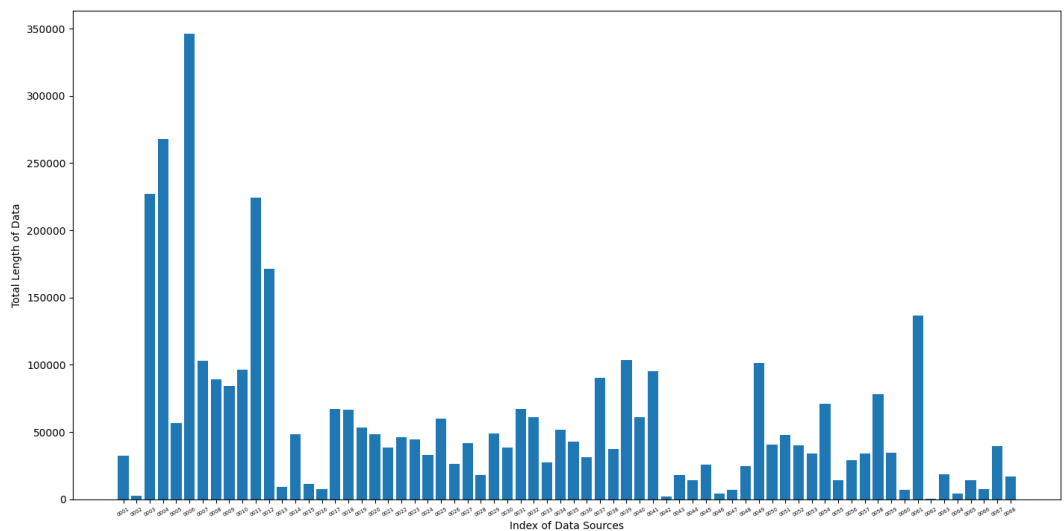

Figure 4. Basic distribution of integrated data set

Based on the aforementioned BIM design quality evaluation methodology, the BIM models submitted in the 68 assignments were scored. The highest score is 90, the lowest is 52, and the mean score is 70.9, with a standard deviation of 7.61. Given that the scores were given manually and may include subjectivity, thereby potentially compromising the reliability of the data. Through statistical analysis, as illustrated in Figure 5, the scores are shown to conform to a normal distribution. This aligns with the typical distribution characteristics expected of scores, thereby affirming the reliability of the data set.

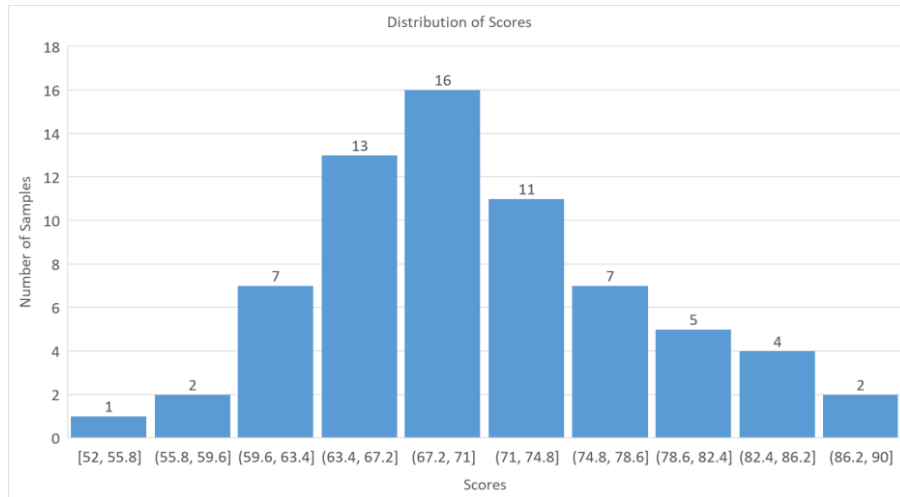

Figure 5. Distribution of scores

## 4.2 Model Comparing and Selection

Prior to conducting the model selection experiments, it is necessary to first select a set of hyperparameter values. Based on the analysis of the integrated data set, the length *N* was initially set to 30,000. The selected behavioral features totaled 29, and following general guidelines, the total number of training samples should be at least 10 times the number of features. Therefore, the total number of samples generated through data augmentation should be at least around 300. Calculations showed that setting *s* to 5,000 resulted in a total of 493 augmented samples, therefore 5,000 is acceptable as the initial value for *s*. In the regression model comparison analysis, all experiments utilized this same set of hyperparameter values.

Among the 42 algorithms tested, 14 models have $R^2$ less than 0 on the test set, and 6 algorithms have $R^2$ greater than 0.3, which are relatively better-performing models. These 6 algorithms all belong to the Ensemble Method category and the $R^2$ of the top five of them on test set are larger than 0.5. This result clearly shows that there exists quantitative relationship between design behaviors and quality, and these models can to some extent learn the relationship between the features and the targeted variable. With a larger sample size generated through data augmentation, machine learning models can portray this relationship even better. Figure 6 presents the 6 best-performing algorithms and their average RMSE and $R^2$ values on the training and test sets, obtained through multiple runs.

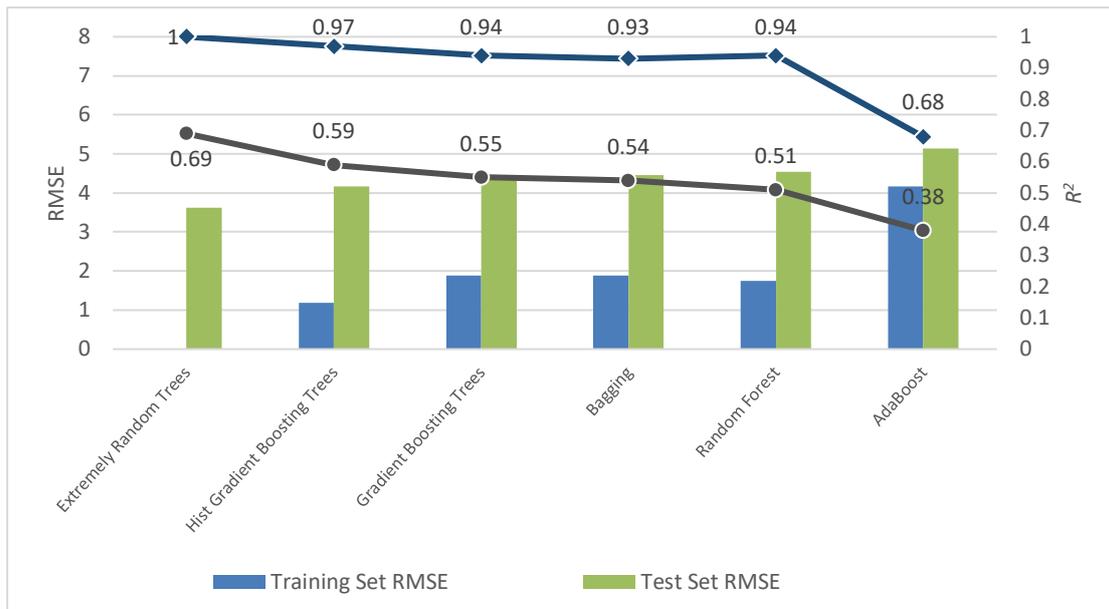

Figure 6. Six best performing models

Among the 6 best-performing algorithms, four of them are tree-based algorithms that achieve improved regression performance through the integration of multiple trees with different additional strategies and methods. In contrast, the Bagging and AdaBoost algorithms do not use fixed type weak regressors, and their core ideas differ from the other four algorithms.

The best-performing Extremely Random Trees (ExtraTrees) algorithm is an ensemble method that constructs multiple decision trees, each trained on a random subset of the features. Unlike traditional decision trees, ExtraTrees introduces additional randomness by splitting nodes using fully random cut-points, rather than searching for the optimal split [36]. The superior performance of the ExtraTrees algorithm in the current study may be attributed to its ability to capture the complex relationships within the data, while maintaining good generalization through the integration of highly randomized decision trees.

The Bagging (Bootstrap Aggregating) algorithm is another ensemble method that generates multiple base models, typically decision trees, by training each on a random bootstrap sample of the original training data. The final prediction is obtained by aggregating the outputs of the individual base models, often through averaging. Bagging can effectively reduce the variance of the base models, leading to improved predictive performance [37].

Considering both the performance of the algorithms and their diversity, the Extremely Random Trees (ExtraTrees) algorithm and the Bagging algorithm were selected as the two optimal algorithms to be used in the subsequent experiments. This choice was made to leverage the strengths of tree-based methods as well as the advantages of the ensemble approach with more diverse base models.

### 4.3 Hyperparameter Optimization Results

Based on the model comparing analysis, the Bagging and ExtraTrees algorithms were selected for further optimization of the hyperparameter values through a controlled experiment.

The experimental results with a fixed *N=30000* and varying *s* are shown in Figure 7. The results indicate that as the step size *s* increases, the number of samples is significantly reduced, which has a substantial impact on the performance of the model training. When *s* is set to a large value of 15,000, the number of samples is only 190, and in this case, both the Bagging and ExtraTrees algorithms struggle to achieve good performance. On the other hand, increasing the value of *s* leads to an improvement in the model's performance. However, when *s* is set to a very small value (e.g., 500), the sample size becomes quite large, which has a limited impact on further improving the model's performance, while also potentially causing information leakage due to the high similarity between samples. This may result in an apparently better model performance but a lower model reliability.

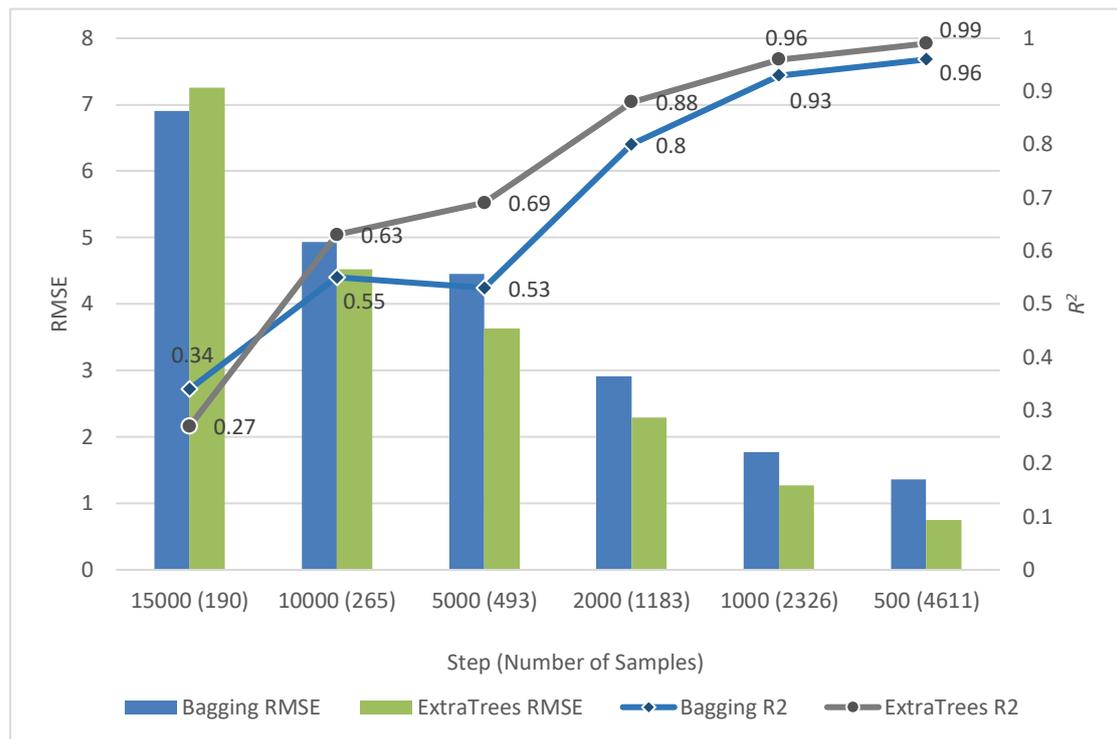

Figure 7. Experiment results of fixed *N=30000* and varying *s*

The experimental results with a fixed step size *s=1000* and varying *N* are shown in Figure 8. By controlling the value of the step size *s*, the sample size can be maintained at a relatively stable level without encountering the issue of insufficient samples.

When *N* is reduced, the sample size increases, but the performance of both the Bagging and ExtraTrees algorithms declines. This may be due to the fact that each sample becomes shorter, resulting in insufficient information content and increased sensitivity to extreme cases, leading to greater differences between the data samples. On the other hand, when *N* is set to a larger value, the performance is already satisfactory, and further improvement becomes less prominent.

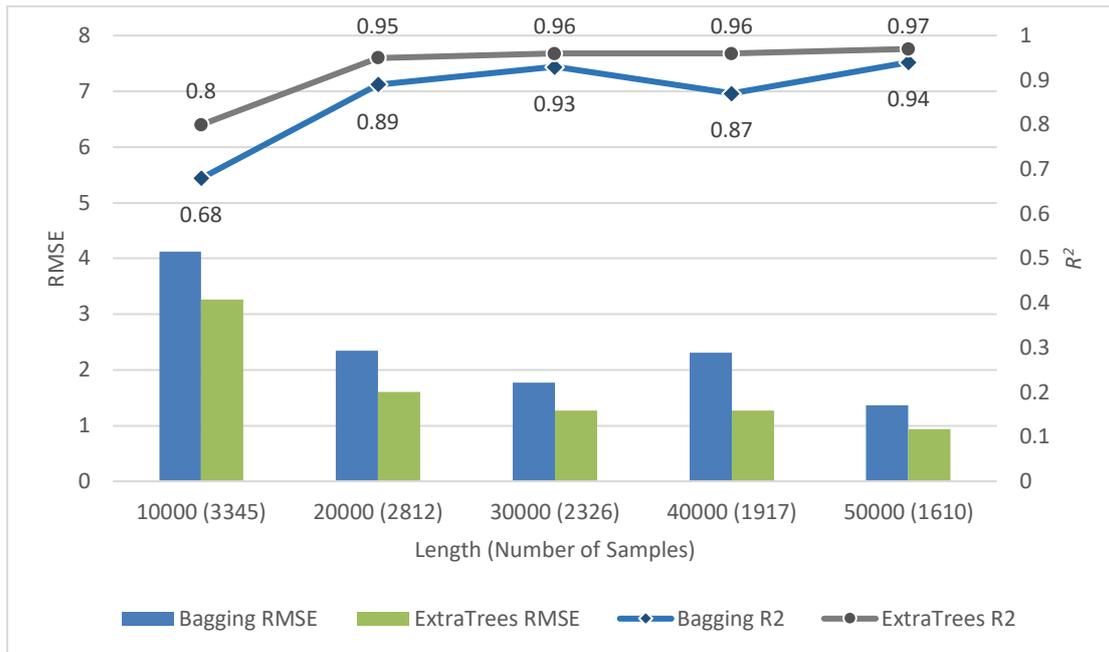

Figure 8. Experiment results of fixed *s=1000* and varying *N*

Based on the comprehensive analysis of the experimental results, the following insights can be drawn. When the value of *N* is fixed, a smaller step size *s* is generally preferred, as it can lead to better model performance. However, setting *s* too small may potentially compromise the reliability of the model. Conversely, when the step size *s* is kept constant, increasing the value of *N* leads to improved model performance. However, there are diminishing returns beyond a certain value of *N*, as the marginal gains become less significant. Throughout the experiments, the ExtraTrees algorithm consistently outperformed the Bagging model in most cases.

Considering the trade-offs, a suitable value for the hyperparameter *N* is 30,000, as it can provide good model performance without excessively large sample size. For the step size *s*, a value of 2,000 seems appropriate, as it can ensure a satisfactory model performance while maintaining a reasonable level of reliability.

With the appropriate hyperparameter settings, the ExtraTrees algorithm-based model achieved a RMSE of 2.29 on the test set, which is significantly smaller than the standard deviation of 7.61 for the scores in the integrated data set. Additionally, the model attained an $R^2$ value of 0.88, indicating its strong explanatory power. The above results further proves that there is a clear quantitative relationship between design behavior and design quality. The machine learning model employed in this study is capable of learning the relationship between the sample features and their corresponding scores, and can provide reliable score estimates for the test samples.

## 4.4 Model Interpretation

Based on the experiment results, this study conducted a model interpretation analysis focusing on the ExtraTrees algorithm with the optimized hyperparameter values. The analysis involved calculating the SHAP values for all features across all samples, and then aggregating the SHAP values for each feature to assess its impact on the model's output.

Figure 9 depicts the absolute SHAP values for each feature. The vertical axis represents the 14

most influential features, ranked in descending order based on the average absolute SHAP value. The final row in the plot corresponds to the aggregated SHAP values for the remaining 15 features.

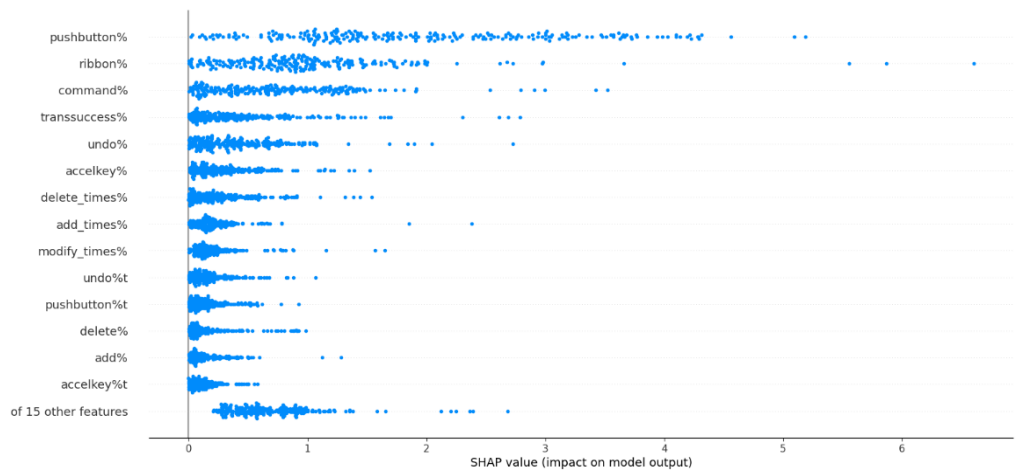

Figure 9. Absolute SHAP value for all features

Overall, the data density features have a greater impact on the model compared to the time density features. The sequential nature of the data may not be effectively reflected through statistical methods, which lead to the features used in this study having difficulty capturing more information along the time dimension. The features ranked at the top are mainly related to command execution, while features related to model components manipulation have relatively less influence. This suggests that the design behavior of executing commands is of great significance for analyzing the characteristics of the designer, as it may reflect the designer's focus, proficiency, and other factors, which ultimately impact the final design quality.

The execution methods of different design commands may have varying degrees of impact on design quality, due to their differing usage frequencies. For example, during the execution of most commands, interactions with dialog boxes, often involving button clicks, are common. This feature is highly correlated with the designer's command execution behavior, and the pushbutton% feature derived from this data can have a more pronounced impact on design quality. In contrast, the usage of keyboard shortcuts in Revit software is not as prevalent and convenient as in CAD software for 2D design. Only a few common shortcuts are typically used, and keyboard shortcuts are not frequently employed during the execution of most commands. Consequently, the data related to accelerator key usage (accelkey%) is not strongly associated with the commands, and thus, it has a relatively smaller influence on design quality.

Figure 10 shows the SHAP values for the corresponding features of each sample. The SHAP values can be positive or negative, and the color of the points reflects the magnitude of the feature values. Redder points indicate larger feature values, while bluer points indicate smaller values. The sign of the SHAP values (positive or negative) reflects the directionality of the relationship between the feature and the model output, i.e., whether the feature is positively or negatively correlated with the model prediction. If the distribution of blue and red points for a particular feature exhibits clear separation or distinction, it can provide important insights. By combining the analysis of data features with the potential scenarios that may be encountered in the design process in practice, abstract data features can be transformed into implications that possess certain practical guiding significance. Based on the most influential features obtained from model analysis, the following two implications can be drawn.

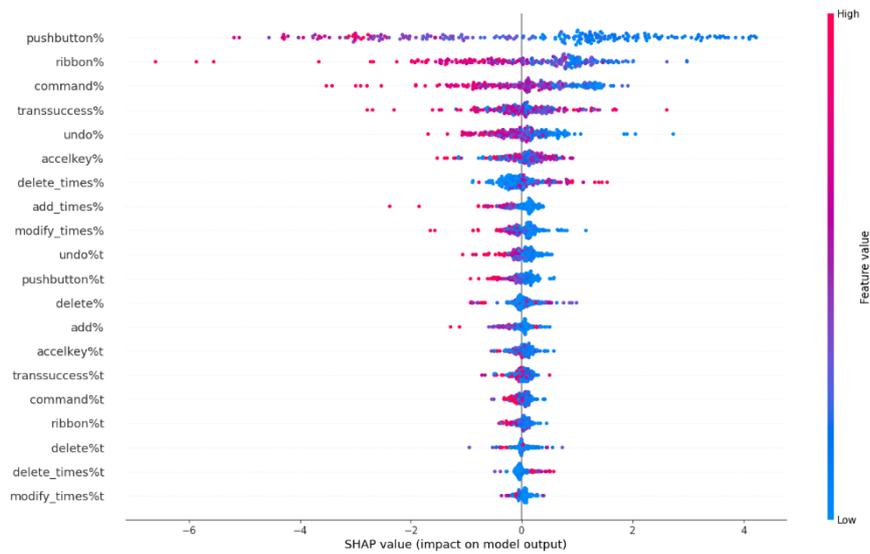

Figure 10. SHAP value for the most influential 20 features

**1）Enhancing designers' design proficiency not only improves design efficiency but also enhances design quality.**

For the majority of samples, the pushbutton% feature values are relatively low, yet they exhibit a positive correlation with the model output. Moreover, a considerable number of samples have SHAP values for this feature reaching as high as 4. This suggests that the student population, as a whole, lacks proficiency in command execution, leading to a smaller proportion of command execution operations in the sample data. Improving proficiency and increasing the efficiency of command execution can significantly enhance the quality of the designs.

For the ribbon% feature, there are more samples with evidently higher feature values compared to the pushbutton% feature. For these samples with elevated ribbon% values, the feature is generally negatively correlated with the model output. As novice users, students are not familiar with methods of command execution, such as keyboard shortcuts, and tend to rely more on explicitly clicking the command buttons in the ribbon menu bar to execute commands. This further indicates that learning and utilizing keyboard shortcuts, while reducing the dependence on the ribbon menu buttons, can improve modeling efficiency. In engineering practice, there is a significant difference in the efficiency of command execution among designers with different skill levels. Efficient command execution relies on the use of shortcuts, which reduces the reliance on mouse clicks and thereby simplifies the execution process of certain commands [26]. In this study, the features related to command execution methods demonstrate that enhancing modeling skills and reducing the frequency of manual key presses not only reflect an improvement in modeling efficiency but also manifest in the enhancement of model quality. There exists a simultaneity between the improvement in modeling efficiency and quality.

**2）Mature design intentions before the start of the design process can reduce rework and improve design quality.**

The three features - delete_times%, add_times%, and modify_times% - also exhibit some regularities. These three features show that the majority of samples are concentrated at relatively low values. However, the density of addition and modification times is positively correlated with modeling quality, while the density of deletion times is negatively correlated. A considerable amount of deletions represent significant rollback and rework in modeling progress, which usually imply major changes of design intentions. This data feature suggests that reducing such major changes of design intentions may help improving design quality. In engineering practice, changes of design intentions may be due to many different reasons. Designers may initially lack a sufficiently clear understanding of the design intentions, leading them to work along a misguided path. Sudden changes of requirement during the design process, or a shift in the designer's own thinking, can also lead to major changes of intentions. All these causes suggest the importance of mature and stable design intentions. Prior to the design process, it is imperative for the designer to thoroughly comprehend the design requirements and to establish relatively mature design intentions after careful consideration. Maintaining the stability of the design intentions and minimizing the occurrence of significant changes can not only ensure the overall efficiency but also help ensure design quality.

In addition, there are some data features that cannot yet be well explained. The differences in the distribution of different-colored points for the transsuccess% and accelkey% features are not very obvious. This may be because the current study did not further subdivide the submitted transactions and the specific functionality of the shortcuts, which may have resulted in the inclusion of some actual ineffective transactions or shortcuts as noise data, weakening the regularity of the feature's impact on the model. For example, designers may unconsciously develop a habit of frequently pressing the "Esc" key, which will also be recorded as a "cancel" shortcut and included in the statistics, but in most cases, pressing the "Esc" key does not have any actual meaning or effect, and does not reflect the behavioral characteristics.

## 5 Discussion

Design behavior data and design result quality are logically intertwined, and there have been numerous related studies in many other fields. However, in the AEC domain, there is a lack of research directly associating design behaviors with design quality.

This study used machine learning methods to directly establish a quantitative relationship between design behavior and design quality. We proposed corresponding solutions to various issues, such as the source of design behavior data, data preprocessing methods and design quality evaluation methods. Finally, quantitative models were established proving a clear relationship between design behavior and design quality, and interpretation analysis of the best-performing model was conducted, leading to a number of further insights.

The proposed method employed in this study has the potential for broader application beyond the specific context investigated. The quantitative modeling and interpretation analysis methods can be adapted and extended to address other relevant issues in this professional field. For example, the relationship between design behavior data and design efficiency, or the relationship between construction process and the quality of finished building. These topics can be investigated by collecting user behavior data from relevant software and conducting quantitative modeling to explain and predict the outcomes of the behaviors. Furthermore, the quantitative model proposed in this study reveals how specific design behaviors impact design quality, which can provide a solid foundation for further research in the field of design process management focusing on how to actively monitor and adjust design quality and efficiency.

The key features identified through the use of model interpretation method for machine learning models, as well as the ways in which these features influence the model outputs, can help us understand how designers' behaviors impact design quality. By considering the practical challenges that designers may face, we can link high deletion density to major changes of design intentions, and associate the frequency of shortcut and button clicking behaviors with designers' proficiency. This reveals the connection between data-driven conclusions and real-world practical issues. However, this analysis cannot explain some unexpected results and these conclusions require further qualitative research, such as interviews, questionnaires, and surveys, to effectively supplement and support findings related to designers. In the future, qualitative studies focusing on designers' behaviors could provide valuable new insights for this knowledge domain, deepening our understanding of the intrinsic mechanisms that determine outcomes in complex design processes.

This study has some limitations and issues that require further studies:

1) The data acquired is insufficient in volume, making it challenging to support the training of more complex deep learning algorithms with higher data requirements.

2) The designers studied were students, whose modeling capabilities may not be representative of the industry standard. More diverse data samples are needed to improve the accuracy of the models in real-world scenarios.

3) Design behavior data has a strong sequential characteristic, which the statistical features may have difficulty capturing. Exploring sequential analysis methods, such as Recurrent Neural Network and Long Short-Term Memory Network, could be a valuable direction for further research.

# 6 Conclusion

Regarding design quality management issues in the AEC industry, there exist a research gap on how design behaviors impact design quality. To identify and quantitatively describe the relationship between design behavior and design quality, this study proposed a novel approach integrating of log data and real-time data collection to acquire BIM design behavior data, as well as a quantitative evaluation system for BIM modeling design quality. Feature engineering and machine learning model interpretation were leveraged to establish quantitative regression models, and identify key features and behavioral characteristics that have a significant impact on design quality.

The key conclusions of this study are threefold. First, this study confirmed a quantifiable relationship between design behaviors and design quality outcomes for the first time, addressing the mentioned research gap. This contributes to the body of knowledge by transforming the nebulous logical connection into concrete, measurable models.

Second, machine learning models can accurately describe these relationships. The best-performing machine learning model constructed using the Extremely Random Trees algorithm was able to assess design quality based on the design behavior features, achieving an $R^2$ value of 0.88 on the test set, demonstrating the model's strong explanatory power. This modeling capability provides a solid foundation for future research on design quality management process.

Third, our analysis identified the critical design behavior features that most significantly impact final design quality. Through interpretation analysis of the model, the data density features related to the execution of design commands and manipulation of model components were found to have the most important influence on design quality. This analysis also empirically supports the notion that improving designers' skill levels, as well as ensuring more mature design intentions prior to the design process, can effectively reduce rework and improve overall design quality. These insights deepen our understanding of the underlying reasons for varying levels of design quality in the design workflow, and offer guidance for ensuring design quality.

Future research may focus on qualitative analysis on the topic, exploring additional algorithms and techniques to further enhance the predictive power of the models, potentially considering the sequential nature of design behaviors. Larger and more diverse datasets would support the utilization of even more sophisticated models.

# Acknowledgement

The authors are grateful for the financial support received from the National Natural Science Foundation of China (No. 52378306).